\documentclass[10pt,twocolumn,letterpaper]{article}

 \usepackage[pagenumbers]{wacv} 

\definecolor{wacvblue}{rgb}{0.21,0.49,0.74}
\usepackage[pagebackref,breaklinks,colorlinks,allcolors=wacvblue]{hyperref}
\usepackage{standalone}
\usepackage{tikz}
\usepackage{graphicx}
\usepackage{cuted}
\usepackage{bm}
\usepackage{anyfontsize}
\usepackage{placeins}
\usepackage[abbreviations,toc]{glossaries-extra}
\usepackage[accsupp]{axessibility}  
\usetikzlibrary{arrows.meta, calc, positioning, shapes.geometric, shadows.blur, fit, backgrounds}

\glsdisablehyper

\def\wacvPaperID{1498} 
\def\confName{WACV}
\def\confYear{2027}

\newabbreviation{okls}{OKLS}{Object Keyline Similarity}
\newabbreviation{oks}{OKS}{Object Keypoint Similarity}
\newabbreviation{sap}{sAP}{Structural Average Precision}
\newabbreviation{pnp}{PnP}{Perspective-$n$-Point}
\newabbreviation{pnpl}{PnPL}{Perspective-$n$-Point-and-Line}
\newabbreviation{nms}{NMS}{non-maximum suppression}

\title{Keyline Detection with YOLO:\\ Runway Perception for Vision-Based Aircraft Landing}

\author{
{\fontsize{12pt}{14.4pt}\selectfont Leander Fritsch}$\,^{1,2,3}$ \qquad 
{\fontsize{12pt}{14.4pt}\selectfont David R{\"u}gamer}$\,^{2,3}$ \qquad 
{\fontsize{12pt}{14.4pt}\selectfont Alexej Klushyn}$\,^{1}$\\[2mm]
{\small $^1$\textit{Airbus Central Research \& Technology}}
\quad
{\small $^2$\textit{LMU Munich}}
\quad
{\small $^3$\textit{Munich Center for Machine Learning (MCML)}}
}

\begin{document}
\maketitle
\begingroup
\begin{NoHyper}
\renewcommand\thefootnote{}
\footnotetext{Corresponding author: \texttt{leander.fritsch@airbus.com}}
\end{NoHyper}
\endgroup
\begin{abstract}
Modern object detectors can predict keypoints as per-object geometric features, but keypoints are inherently vulnerable to occlusion and truncation. 
Object edges can provide complementary information, often remaining observable even when partially occluded or when their defining endpoints leave the field of view.
We therefore introduce object keyline detection, the task of predicting a fixed, semantically ordered set of infinite lines per object instance. 
Unlike the outputs of general line-segment detectors, each keyline is associated with an object instance and identified by a semantic index, providing a line-level analog of keypoints.
To solve this task, we extend YOLO with keyline prediction parameterized in continuous Hesse-normal form $[\rho,\cos\theta,\sin\theta]$, and introduce \gls{okls}, which serves as an evaluation metric, target assignment signal, and regression loss.
We instantiate this formulation on runway perception in vision-based landing, a motivating example containing semantic linear features that remain partially observable throughout the approach.
Experiments show that our representation achieves the highest detection accuracy among four evaluated line parameterizations, and keylines integrate directly into standard YOLO heads as additional output channels without degrading keypoint performance or adding meaningful latency.
\end{abstract}    
\glsresetall
\section{Introduction}
\label{sec: introduction}

\begin{figure}[t]
  \centering
  \includegraphics[width=\linewidth]{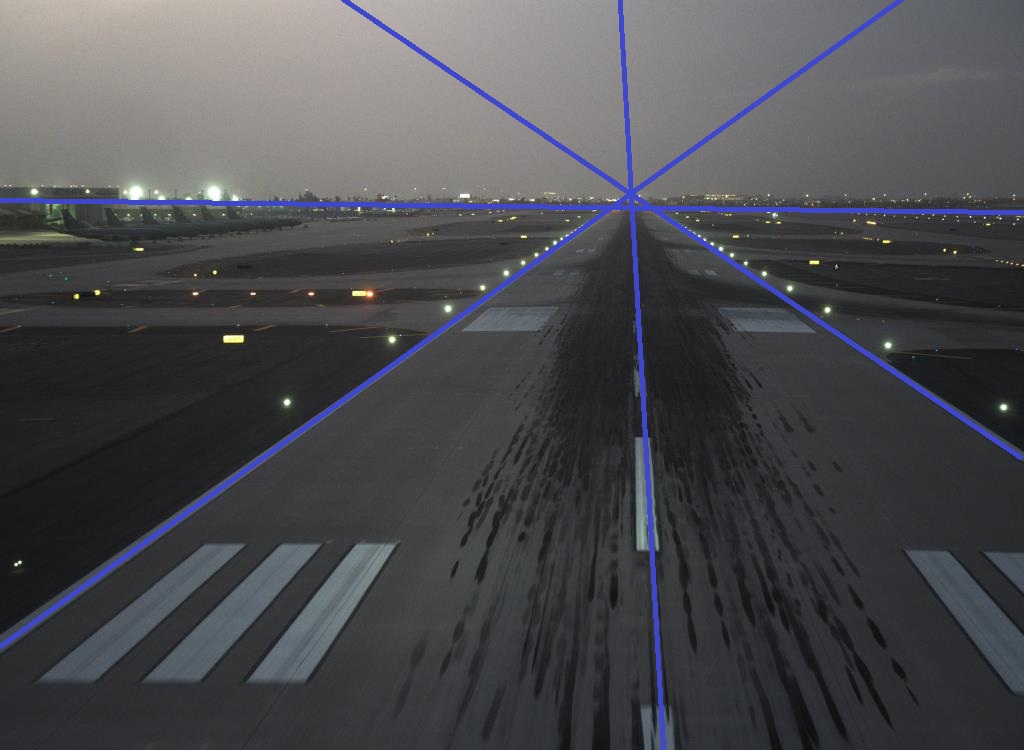}
  \caption{A partially visible runway during final approach. The two near corners are truncated by the image boundary, preventing keypoint-based pose estimation, while four of five runway keylines remain observable, providing line constraints for pose estimation.}
  \label{fig:partially_visible_runway}
\end{figure}

Keypoint prediction has become a widespread capability of modern object detectors, as the detection of semantic per-object features enables downstream geometric tasks like pose estimation. 
Keypoints, however, can only represent point-based features. 
For objects characterized by prominent edges, this is limiting: a keypoint that is occluded or truncated provides no information, whereas the corresponding edge can remain partially observable and still convey orientation and perspective (\eg,~\cref{fig:partially_visible_runway}).

General line-segment detectors address the extraction of linear image structures, but they solve a different problem.
Methods such as M-LSD~\citep{gu2022towards}, HAWP~\citep{xue2020holistically}, and LINEA~\citep{janampa2025linea} predict unordered, image-level collections of finite line segments.
Their outputs are neither associated with individual object instances nor assigned a stable semantic identity. 
Conversely, object detectors can predict bounding boxes and semantic keypoints for multiple object instances, but do not include line features among their per-object predictions.
Thus, existing approaches do not provide a line-level analog of semantic keypoint detection.

\begin{figure*} 
    \centering
    \resizebox{0.85\textwidth}{!}{%
        \input{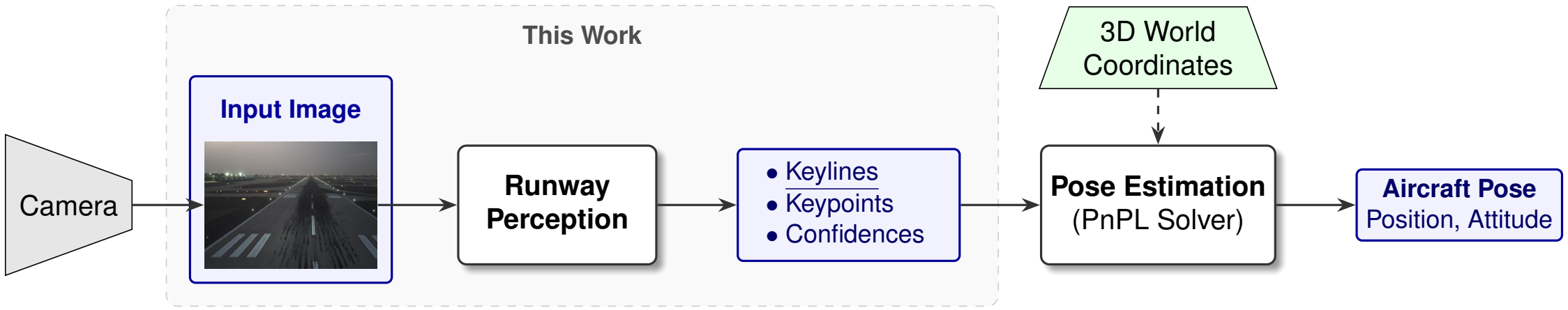}
    }
    \caption{The basic components of aircraft pose estimation for vision-based landing. An object detection algorithm detects a set of runway objects and their predetermined features, while a subsequent pose estimation algorithm estimates the aircraft's pose. In this work we extend the detected feature set by keylines and evaluate them in the image domain.}
    \label{fig:VLS}
\end{figure*}

To bridge this gap, we introduce \emph{object keyline detection}, extending object detectors to predict instance-bound geometric lines alongside keypoints. 
Unlike finite line segments, keylines are predicted as infinite lines without spatial extent targets, while finite support annotations define where line accuracy is evaluated. This formulation provides a direct line-level analog to semantic keypoint detection.

We instantiate object keyline detection for runway perception in vision-based aircraft landing \citep{balduzzi2021neural} (see \cref{fig:VLS}).
In such a system, correct estimation of runway geometry throughout the final approach is critical.
However, corner keypoints frequently suffer from field-of-view truncation during the close-range phase (\cref{fig:partially_visible_runway}), causing \gls{pnp} \citep{lepetit2009ep} solvers to lose geometric constraints while edge features remain observable.
Keylines provide a persistent geometric representation under these conditions and could serve as a fallback when keypoint detection fails.
Since recent runway detection systems commonly employ YOLO \citep{redmon2016you} models \citep{li2024yolo,dai2024yomo}, we extend YOLO keypoint detection \citep{maji2022yolo} with keyline prediction. 

Object keyline detection raises two central questions: how should keylines be parameterized for stable regression in YOLO models, and how should they be compared during model evaluation and optimization?
We address the first through a controlled comparison of line parameterizations and the second by introducing \emph{Object Keyline Similarity} (\glsxtrshort{okls}), an adaptation of \gls{oks} to keylines. \glsunset{okls} 
We make the following contributions: 
\begin{itemize}
    \item We formulate object keyline detection as predicting a fixed, semantically ordered tuple of infinite lines per object instance, instantiating it as a native YOLO extension and demonstrating it on runway perception.
    \item We introduce \gls{okls} (Object Keyline Similarity) as a unified evaluation metric, target assigner, and regression loss, enabling AP and AR metrics for keylines.
    \item We benchmark four line parameterizations and find the continuous Hesse normal form $(\rho, \cos\theta, \sin\theta)$ to be the most accurate of the evaluated representations on our benchmark.
    \item We evaluate fused- and split-head architectures for joint keypoint--keyline prediction, demonstrating that a fused head adds keylines without degrading keypoint accuracy or adding meaningful latency.
    \item We compare against general line-segment detectors under an oracle matching protocol, demonstrating higher average precision in detecting object-associated lines.
\end{itemize}
\section{Related work}
\label{sec:related_work}

\subsection{Vision-based runway perception}
\label{sec:vbl}

Early runway perception focused on aerial mapping via hand-crafted edge finders and heuristic filtering \citep{huertas1990detecting}. Several recent runway-perception systems employ YOLO detectors to localize runways in approach imagery \citep{dai2024yomo, li2024yolo} and extract keypoints (\eg, corners or markings) to solve for camera pose via \gls{pnp} ~\citep{ferreira2025vision, yu2025vi}.
However, corner truncation or occlusion during final approach causes keypoint-only pose solvers to lose geometric redundancy or fail entirely \citep{balduzzi2021neural}. 
To address this, \citet{balduzzi2021neural} proposed a two-stage cropping network that regresses a global runway-geometry vector, including a vanishing point and line angles.
\citet{sun2026visual} present a hybrid approach, combining YOLO region proposals with an unstructured classical line-segment detector. 
In contrast, our formulation provides a single-stage pipeline that predicts semantically defined keylines to support downstream \gls{pnpl} solvers \citep{vakhitov2016accurate}, although evaluating its impact on aircraft pose accuracy is beyond the scope of this work.

\subsection{Line detection}
\label{sec:lsd}

Classical Hough transforms \citep{duda1972use} and modern learned variants \citep{lee2017semantic,jin2020semantic,zhao2021deep,jin2021harmonious} predict infinite line parameters but focus on prominent, image-spanning lines.
Conversely, line-segment detectors \citep{huang2020tp,gu2022towards, xue2019learning, xue2020holistically, xu2021line, janampa2025linea} frequently predict shorter edges, but rely on the detection of segment endpoints.
Crucially, both categories lack modeling of object-association or semantic line identities.
A central challenge in line detection is selecting an effective geometric line parameterization. 
Existing formulations include polar coordinates $(\rho, \theta)$ \citep{zhao2021deep,jin2021harmonious}, 2D endpoints \citep{xu2021line,janampa2025linea}, attraction vectors \citep{xue2019learning,xue2020holistically}, tri-point center offsets \citep{huang2020tp,gu2022towards}, distance and orientation fields \citep{pautrat2023deeplsd}, or row-wise offsets \citep{Honda_2024_WACV}. 
Because these representations were designed for varying models and a structurally different task, their performance does not directly transfer to instance-centric keyline regression in YOLO, motivating our systematic comparison (\cref{sec:alt_rep}).
Likewise, existing metrics (\eg $\text{sAP}$ \citep{zhou2019end} or $\text{HIoU}$ \citep{jin2021harmonious}) are not designed for object keyline detection.
\section{Methodology}
\label{sec: methods}

\subsection{Task formulation}
\label{sec:task}

We formulate \emph{object keyline detection} as an extension of object keypoint detection.
Given an image $I$, the model predicts a variable-size set of $N$ object detections,
\begin{equation*}
    \mathcal{D}(I)=\left\{\left(\hat b_n, \hat c_n, \hat q_n,\hat{\mathcal K}^\mathrm{points}_n, \hat{\mathcal K}^\mathrm{lines}_n\right)\right\}_{n=1}^{N},
    \label{eq:predictions}
\end{equation*}
where $\hat b_n$, $\hat c_n$, and $\hat q_n$ denote the predicted bounding box, category, and object confidence of detection $n$. Each detected object is associated with fixed, semantically ordered tuples of $K$ keypoints $\hat{\mathcal K}^\mathrm{points}_n = ((\hat{\boldsymbol{y}}_{n,k},\hat{v}_{n,k}^{\mathrm{kp}}))_{k=1}^{K}$ and $L$ keylines 
\begin{equation*} 
    \hat{\mathcal{K}}^\mathrm{lines}_n = \left(\left(\hat{\boldsymbol{l}}_{n,l},\hat{v}_{n,l}^{\mathrm{kl}}\right) \right)_{l=1}^{L},
\end{equation*}
where $\hat{\boldsymbol{l}}_{n,l}$ denotes the geometric prediction of keyline $l$, and $\hat{v}_{n,l}^{\mathrm{kl}} \in [0,1]$ denotes the corresponding predicted visibility score. 
The model therefore always produces $L$ keyline predictions for each object, but may indicate that certain associated object edges are out of view by predicting a low visibility score.
A predicted keyline $\hat{\boldsymbol{l}}_{n,l}$ specifies an infinite line in the 2D image space.
Each ground-truth keyline tuple 
\begin{equation*} 
    \mathcal{K}^\mathrm{lines}_n = \left(\left(\boldsymbol{e}_{n,l,1}, \boldsymbol{e}_{n,l,2}, v_{n,l}^{\mathrm{kl}}\right) \right)_{l=1}^{L}
\end{equation*}
is annotated with the same fixed semantic ordering, which directly defines the correspondence between predicted keylines and annotations.
Let $\Omega = [0, W] \times [0, H] \subset \mathbb{R}^2$ denote the image canvas.
An annotation $l$ consists of two support endpoints $\boldsymbol{e}_{n,l,1}, \boldsymbol{e}_{n,l,2} \in \Omega$, specifying the finite support over which line accuracy is evaluated and a visibility label $v_{n,l}^{\mathrm{kl}}$, specifying whether that line is visible ($v_{n,l}^{\mathrm{kl}} > 0$) or not ($v_{n,l}^{\mathrm{kl}} = 0$).
For edges extending beyond the image boundaries, the support segment is clipped to the image boundary $\partial \Omega$. 
Annotated object edges that are completely outside of $\Omega$ receive visibility label $v_{n,l}^{\mathrm{kl}}=0$, while others receive a positive visibility label. 

\subsection{Keyline representation and decoding}
\label{sec:representation}

We adapt dense object detectors to keyline prediction using a cell-relative parameterization and decoding scheme.

\paragraph{Global representation.}
We parameterize each predicted keyline in Hesse normal form as
\begin{equation*}
    \boldsymbol{\hat l}_l =
    [\hat \rho_l,\hat n_{x,l},\hat n_{y,l}],
    \qquad
    \hat{\boldsymbol{n}}_l =
    \begin{bmatrix}
        \hat n_{x,l}\\
        \hat n_{y,l}
    \end{bmatrix},
    \qquad
    \|\hat{\boldsymbol{n}}_l\|_2 = 1,
    \label{eq: Hesse-normal}
\end{equation*}
where $\hat{\boldsymbol{n}}_l$ is the unit normal vector of the line, and $\hat \rho_l$ is its signed perpendicular distance from the coordinate origin.
A point $\mathbf x = [x, y]^\top$ in model-input coordinates belongs to that line if and only if $\hat{\boldsymbol{n}}_l^\top\boldsymbol{x}=\hat \rho_l$.

We predict the unit normal vector $\hat{\boldsymbol{n}}_l = [\cos\hat{\theta}_l, \sin\hat{\theta}_l]^\top$ rather than the scalar orientation angle $\hat{\theta}_l$, as that avoids the angular wrap-around discontinuity of scalar-angle regression at $\pm\pi$ \citep{zhou2019continuity}. 
With a signed distance $\hat \rho_l \in \mathbb{R}$ and an unrestricted orientation, this representation is continuous and singularity-free.
However, this choice makes it antipodally ambiguous, since $(\hat{\rho},\hat{\boldsymbol{n}})$ and $(-\hat{\rho},-\hat{\boldsymbol{n}})$ describe the same undirected line. 
\gls{okls} (\cref{sec:okls}) addresses this ambiguity.

\paragraph{Cell-relative decoding.} 
At each output grid cell with center $\mathbf{c}_h$ and stride $s_h$, the network predicts unconstrained raw parameters $\tilde{\boldsymbol{l}}_l = [\tilde{\rho}_l, \tilde{n}_{x,l}, \tilde{n}_{y,l}] \in \mathbb{R}^3$.
Directional outputs are passed through $\tanh$ and normalized to obtain the unit normal $\hat{\boldsymbol{n}}_l$.
Because $\tilde{\rho}_l$ is predicted relative to the cell center, the line satisfies $\hat{\boldsymbol{n}}_l^\top(\mathbf{x} - \mathbf{c}_h) = s_h \tilde{\rho}_l$.
Rearranging gives the global Hesse-normal distance $\hat{\rho}_l$ relative to the image origin:
$\hat{\rho}_l = \mathbf{c}_h^\top \hat{\boldsymbol{n}}_l + s_h \tilde{\rho}_l$, yielding the final decoded output $\hat{\boldsymbol{l}}_l = [\hat{\rho}_l, \hat{n}_{x,l}, \hat{n}_{y,l}]$ (detailed derivations in \cref{supp:cell_relative_decoding}).

\subsection{Object Keyline Similarity}
\label{sec:okls}

\paragraph{Background.}
\gls{oks} measures the localization agreement between two keypoint tuples while accounting for object scale, per-keypoint tolerances, and keypoint visibility. 
For predicted keypoints $\hat{\boldsymbol{y}}$ and ground-truth keypoints $\boldsymbol{y}$, it is defined as
\begin{equation*}
    \operatorname{OKS}
    =
    \frac{1}{|\mathcal V^{\mathrm{kp}}|}
    \sum_{k\in\mathcal V^{\mathrm{kp}}}
    \exp\left(
        -
        \frac{
            \|\hat{\boldsymbol y}_k-\boldsymbol y_k\|_2^2
        }{
            2s^2(\kappa_k^{\mathrm{kp}})^2
        }
    \right),
    \label{eq:oks}
\end{equation*}
where $s^2$ denotes the object area, $\kappa_k^{\mathrm{kp}}$ is a per-keypoint hyperparameter controlling the scale-invariant localization tolerance of the $k$-th keypoint, and 
\begin{equation*}
    \mathcal V^{\mathrm{kp}} =\left\{ k\in\{1,\ldots,K\} \,\middle|\, v_k^{\mathrm{kp}}>0 \right\}
\end{equation*}
is the index set of labeled keypoints, ensuring that keypoints outside the image do not contribute to the score.
YOLO keypoint detectors typically adopt \gls{oks} for model evaluation, instance matching, and loss optimization~\citep{maji2022yolo}.

\begin{figure}
    \centering
    \begin{tikzpicture}[
    gt/.style={very thick, green!60!black}, 
    pred/.style={very thick, blue, -{Latex[length=2mm]}},
    norm/.style={thick, orange, -{Latex[length=2mm]}},
    predline/.style={thick, black}, 
    proj/.style={dashed, gray, thick},
    penalty/.style={thick, dashed, gray}
]

\coordinate (GT1_L) at (0, 1.5);
\coordinate (GT2_L) at (3, 2.5);
\draw[gt] (GT1_L) -- (GT2_L);
\fill[green!60!black] (GT1_L) circle (2pt) node[left] {$\mathbf{e}_1$};
\fill[green!60!black] (GT2_L) circle (2pt) node[right] {$\mathbf{e}_2$};

\coordinate (P1_L) at (0.3, 0.5);
\coordinate (P2_L) at (3.3, 1.5);

\draw[predline] ($(P1_L)!-0.3!(P2_L)$) -- ($(P1_L)!1.3!(P2_L)$) node[right] {$\hat{\boldsymbol{l}}$};

\coordinate (Proj1_L) at ($(P1_L)!(GT1_L)!(P2_L)$);
\coordinate (Proj2_L) at ($(P1_L)!(GT2_L)!(P2_L)$);
\draw[proj] (GT1_L) -- (Proj1_L) node[midway, left] {$p_1$}; 
\draw[proj] (GT2_L) -- (Proj2_L) node[midway, right] {$p_2$}; 

\end{tikzpicture}
    \caption{Our proposed \glsxtrfull{okls} uses the orthogonal distances $p_1$ and $p_2$ from ground-truth support endpoints $(\mathbf{e}_1, \mathbf{e}_2)$ to the predicted line $\hat{\boldsymbol{l}}$ as its residual.}
    \label{fig:line_oks}
\end{figure}
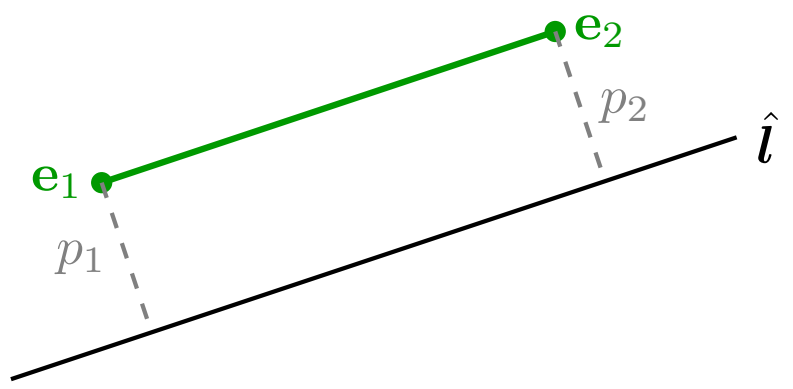

\paragraph{Object Keyline Similarity.}
\gls{oks} is defined for object keypoints, while existing line metrics evaluate lines or line segments without considering object association (cf. \cref{sec:lsd}).
We construct \gls{okls} by replacing the point-to-point residual in \gls{oks} with a line-to-support residual.
Given the predicted infinite line
$\hat{\boldsymbol{l}}_l = [\hat\rho_l,\hat n_{x,l},\hat n_{y,l}]$, 
we compute the signed perpendicular distances to the support endpoints $\boldsymbol{e}_{l,1}$ and $\boldsymbol{e}_{l,2}$ via
\begin{equation*}
    p_{l,j}
    =
    \hat{\boldsymbol{n}}_l^\top\boldsymbol{e}_{l,j}
    -\hat\rho_l,
    \qquad
    \mathbf p_l = \begin{bmatrix}
        p_{l,1} \\ p_{l,2}
    \end{bmatrix},
\end{equation*}
as illustrated in~\cref{fig:line_oks}.
\gls{okls} is then defined as
\begin{equation*}
\text{OKLS} = \frac{1}{|\mathcal V^{\mathrm{kl}}|} \sum_{l \in \mathcal V^{\mathrm{kl}}} \exp \left( -\frac{\|\mathbf p_l\|_2^2}{2 s^2 (\kappa^{\mathrm{kl}})^2} \right),
\end{equation*}
where 
\begin{equation*}
    \mathcal V^{\mathrm{kl}} =\left\{ l\in\{1,\ldots,L\} \,\middle|\, v_l^{\mathrm{kl}}>0 \right\}
\end{equation*}
is the index set of labeled lines and $\kappa^{\mathrm{kl}}$ denotes the scale-invariant localization tolerance, which can optionally be defined per line ($\kappa_l^{kl}$) to reflect differing physical widths or annotation variances.
Although the predicted line is infinite, its finite ground-truth support determines where accuracy is relevant and captures the lever-arm effect of angular error. 
The line-to-support residual inherently treats lines as undirected: $(\hat{\rho},\hat{\boldsymbol{n}})$ and $(-\hat{\rho},-\hat{\boldsymbol{n}})$ yield the same OKLS value.

\paragraph{OKLS for model evaluation.}
For evaluation, we replace \gls{oks} within the standard COCO keypoint evaluation  protocol~\citep{lin2014microsoft} with our \gls{okls} formulation, producing $\mathrm{AP}_{\mathrm{KL}}$ and $\mathrm{AR}_{\mathrm{KL}}$ metrics (where ``KL'' denotes keylines).
In this protocol, \gls{okls} is used in the matching process between predicted and ground-truth instances and categorizes matches as true positive or false positive by thresholding it. 
These metrics are scale-invariant and evaluated over the annotated supports.
For the vision-based landing application, they can be interpreted as a rough proxy metric of keyline usefulness for the downstream aircraft pose estimation, as line-based pose estimation methods, such as \citet{vakhitov2016accurate}, often optimize similar line-to-endpoint distances.
\\
\paragraph{OKLS-based training objective.}
We define the keyline regression loss analogously to the keypoint loss \citep{maji2022yolo} as
\begin{equation*}
    \mathcal{L}_{\text{kl-reg}} = \frac{1}{\sum_{n,m} w_{n,m}} \sum_{n=1}^N \sum_{m=1}^M w_{n,m} \cdot (1 - \text{OKLS}_{n,m}).
    \label{eq:alignment_weighting}
\end{equation*}
$\text{OKLS}_{n,m}$ is the \gls{okls} evaluation between the $n$-th candidate prediction and $m$-th ground-truth object, while the weights $w_{n,m} \in [0, 1]$ are determined by a task-aligned target assigner (\eg, TOOD~\citep{feng2021tood}). It matches ground-truth and predicted objects and assigns weights based on bounding-box overlap, classification confidence and \gls{oks}, which we replace with \gls{okls} to use keylines instead of keypoints in the assignment process.
Predicted keyline visibility $\hat{v}_{n,l}^{\mathrm{kl}}$ is supervised with the keypoint visibility loss. 

This training objective inherits the outlier rejection of \gls{oks}: the gradients vanish for large residuals, so outliers influence the model less than inliers \citep{barron2019general}. For keypoints, this results in more stable training and higher final AP than an $L_1$ loss~\citep{maji2022yolo}.

\paragraph{Multi-task integration.}

For joint training, we replace the standard pose loss with $\mathcal{L}_{\text{pose}} = \alpha \, \mathcal{L}_{\text{kp-reg}} + (1\!-\!\alpha) \, \mathcal{L}_{\text{kl-reg}}$, where $\alpha$ is a hyperparameter.
Similarly, we compute the assigner weights $w_{n,m}$ using the combined similarity $\alpha \, \mathrm{OKS}_{n,m} + (1\!-\!\alpha) \, \mathrm{OKLS}_{n,m}$, so both keypoints and keylines inform target selection.

\subsection{Keyline detection architecture} 
\label{sec:model_architecture} 

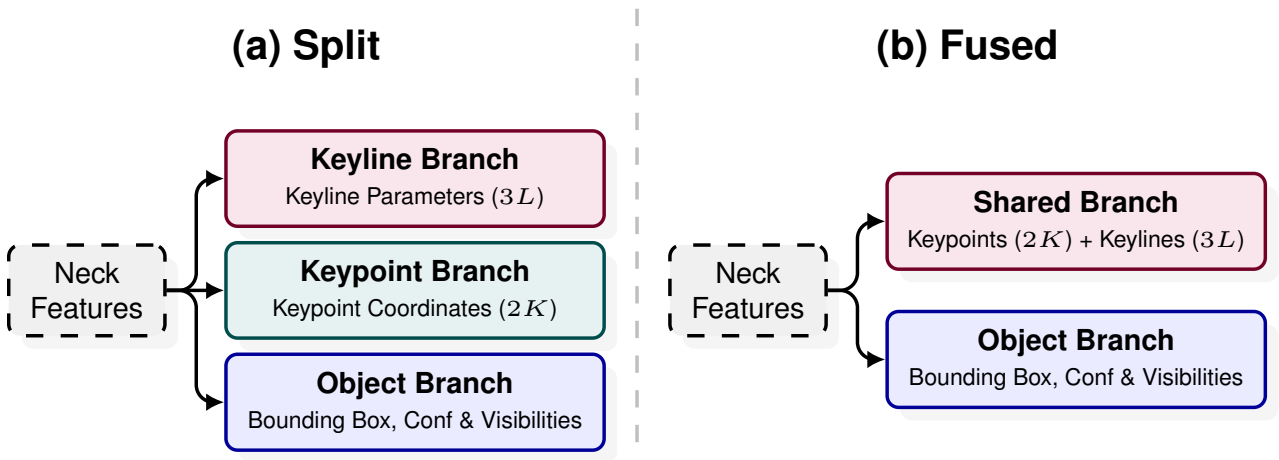
\begin{figure}
  \centering
  \resizebox{\linewidth}{!}{\begin{tikzpicture}[
   node distance=0.25cm and 0.35cm,
   font=\sffamily\tiny,
   >={Latex[width=1.1mm,length=1.1mm]},
   base/.style={
       draw, rounded corners=2pt, align=center, semithick,
       drop shadow={opacity=0.08}, inner sep=3pt, minimum height=0.52cm
   },
   input/.style={
       base, fill=gray!12, dashed, minimum width=0.95cm
   },
   fusedbranch/.style={
       base, fill=purple!10, draw=purple!60!black, minimum width=2.3cm
   },
   kpbranch/.style={
       base, fill=purple!10, draw=purple!60!black, minimum width=2.3cm
   },
   klbranch/.style={
       base, fill=teal!10, draw=teal!60!black, minimum width=2.3cm
   },
   objectbranch/.style={
       base, fill=blue!8, draw=blue!60!black, minimum width=2.3cm
   },
   title/.style={
       font=\sffamily\bfseries\scriptsize, align=center
   }
]
   \node[input] (input_b) {Neck\\Features};
   \node[kpbranch, right=0.35cm of input_b, yshift=0.68cm] (kp_b) {\textbf{Keyline Branch}\\[0pt]\fontsize{4pt}{5pt}\selectfont Keyline Parameters ($3L$)};
   \node[klbranch, right=0.35cm of input_b, yshift=0cm] (kl_b) {\textbf{Keypoint Branch}\\[0pt]\fontsize{4pt}{5pt}\selectfont Keypoint Coordinates ($2K$)};
   \node[objectbranch, right=0.35cm of input_b, yshift=-0.68cm] (obj_b) {\textbf{Object Branch}\\[0pt]\fontsize{4pt}{5pt}\selectfont Bounding Box, Conf \& Visibilities};
   \draw[->, semithick, rounded corners=3pt] (input_b.east) -- ++(0.18,0) |- (kp_b.west);
   \draw[->, semithick] (input_b.east) -- (kl_b.west);
   \draw[->, semithick, rounded corners=3pt] (input_b.east) -- ++(0.18,0) |- (obj_b.west);
   \node[input, right=3.05cm of input_b] (input_a) {Neck\\Features};
   \node[fusedbranch, right=0.35cm of input_a, yshift=0.42cm] (fused_a) {\textbf{Shared Branch}\\[0pt]\fontsize{4pt}{5pt}\selectfont Keypoints ($2K$) + Keylines ($3L$)};
   \node[objectbranch, right=0.35cm of input_a, yshift=-0.42cm] (obj_a) {\textbf{Object Branch}\\[0pt]\fontsize{4pt}{5pt}\selectfont Bounding Box, Conf \& Visibilities};
   \draw[->, semithick, rounded corners=3pt] (input_a.east) -- ++(0.18,0) |- (fused_a.west);
   \draw[->, semithick, rounded corners=3pt] (input_a.east) -- ++(0.18,0) |- (obj_a.west);
   \path (input_b.west) -- (kp_b.east) coordinate[midway] (center_b);
   \node[title, anchor=south] (title_b) at (center_b |- kp_b.north) [yshift=0.25cm] {(a) Split};
   \path (input_a.west) -- (fused_a.east) coordinate[midway] (center_a);
   \node[title, anchor=south] (title_a) at (center_a |- kp_b.north) [yshift=0.25cm] {(b) Fused};
   \path (kp_b.east) -- (input_a.west) coordinate[midway] (divider_x);
   \draw[dashed, gray!50, semithick]
     (divider_x |- title_b.north) -- (divider_x |- obj_b.south);
\end{tikzpicture}}
  \caption{Head architecture variants for joint keypoint and keyline prediction: (a) Split: uses a dedicated keyline branch; (b) Fused: shares a single geometric branch across tasks.}
  \label{fig:head_architectures}
\end{figure}

We extend standard YOLO keypoint detection models (\eg, YOLO-Pose~\citep{maji2022yolo}) to predict keylines alongside bounding boxes and keypoints.
Modern YOLO architectures comprise a backbone, neck and multi-scale prediction heads.
At each grid cell across the output grids, the model produces an object hypothesis (\cref{sec:task}).
To support keyline detection, each hypothesis is extended by $4L$ parameters, encoding the geometric prediction $\hat{\boldsymbol{l}}_l = [\hat{\rho}_l, \hat{n}_{x,l}, \hat{n}_{y,l}]$ and the visibility score $\hat{v}_l^{\mathrm{kl}}$ for each keyline $l\in\{1,\ldots,L\}$.
As illustrated in \cref{fig:head_architectures}, we evaluate two structural integrations for joint keypoint and keyline prediction:

\begin{itemize}
  \item \textbf{Split:} Duplicates the geometric prediction branch to isolate keyline parameter regression from keypoint estimation, while adding $L$ visibility logits to the classification output in the object branch.
  \item \textbf{Fused:} Keypoint and keyline predictions share the existing pose branch, extending only the output layer by $3L$ keyline channels and sharing keyline visibility logits within the object branch as in the split variant.
\end{itemize}
During inference, predicted keyline parameters are mapped back to original image dimensions (\cref{supp:inference}).

\section{Experimental setup}
\label{sec:exp_protocol}

\subsection{Data}
\label{subsec:data}
We train on a curated subset of a proprietary dataset\footnote{We use a subset of the original data for confidentiality reasons.}$^,$\footnote{We do not use the open-source LARD dataset \citep{bougacha2026lard} because it does not contain partially visible runways.} of real approach images, split into training, validation, and test data.
The training set contains 6730 images from 230 approach trajectories at 95 airports.
Images contain an average of 1.49 annotated runways (maximum 4), with an average of 3.4 out of 4 corners (85\%) and 4.7 out of 5 lines (94\%) visible.
To prevent geographic data leakage, airports are strictly disjoint between training, validation, and test data.
The validation (712 images, 40 trajectories) and test sets (683 images, 38 trajectories) cover 10 airports each, disjoint from training, and have similar ratios of truncated images to the training data.
The overall dataset represents a wide range of real-world conditions, including various times of day (and night), weather conditions, and approach distances from -6 km to +1 km relative to the runway threshold. 
The annotation file extends the COCO annotation format with keyline support endpoints and visibility labels.
Input images are scaled and vertically padded to $1024\times1024$ pixels. 
Training augmentations include color and contrast changes, blur, shadows, and coarse dropout. 

\subsection{Metrics}
\label{subsec:Metrics}
Keyline detection is evaluated using $\text{AP}_{\text{KL}}$ and $\text{AR}_{\text{KL}}$ calculated over standard COCO thresholds $\{0.50,0.55,\ldots,0.95\}$ with a fixed localization tolerance $\kappa^{kl} = 0.035$.
While this choice is application-driven rather than derived from variation among repeated independent annotations \citep{lin2014microsoft}, holding $\kappa$ constant provides a fair benchmark for relative model comparisons and is a common practical choice (\eg \citep{fang2022alphapose}). 
Keypoint evaluation uses standard \gls{oks}-based $\text{AP}_{\text{KP}}$ and $\text{AR}_{\text{KP}}$ with an identical tolerance $\kappa_{k}^{kp} = 0.035$.

\paragraph{PnPL constraint availability.}
To quantify whether the detector provides a sufficient number of accurately localized feature correspondences for a generic minimal \gls{pnpl} solver, we introduce PnPL Constraint Availability ($\mathcal{A}_{\text{PnPL}}$). It is defined as the proportion of evaluated images containing at least $N_{\mathrm{min}}=3$ valid feature predictions. A predicted keypoint or keyline is considered valid if its predicted visibility satisfies $\hat{v}_{n,l}^{\mathrm{kl}} \ge \tau_\text{conf}$ or $\hat{v}_{n,k}^{\mathrm{kp}} \ge \tau_\text{conf}$ and the associated localization score (defined as the $k$-th or $l$-th element of \gls{oks} or \gls{okls}) satisfies $\mathrm{OKS} \ge \tau$ or $\mathrm{OKLS} \ge \tau$, respectively. 
We report $\mathcal{A}_{\text{PnPL}}$ for $\tau=\tau_\text{conf}=0.50$.

This metric measures image-domain constraint availability. Meeting the three-correspondence count is a necessary generic minimum but is not sufficient to guarantee successful pose estimation, as the metric does not test geometric degeneracy (\eg, detecting only left, right and centerline), numerical conditioning, or aircraft pose error.

\subsection{Keyline representation ablation}
\label{sec:alt_rep}

We benchmark the continuous Hesse-normal representation ($[\rho, n_x, n_y]$) against three parameterizations adapted from the line segment detection literature (cf. \cref{sec:lsd}) integrated within the same YOLO-NAS architecture:

\begin{itemize}
    \item \textbf{Polar Hesse Normal} $(\rho, \theta)$: Regresses a scalar orientation angle $\theta$ instead of a continuous normal vector. \gls{okls} loss is applied by computing the regular Hesse normal form as $[\rho, \cos \theta, \sin \theta]$.
    \item \textbf{Endpoints ($[x_1, y_1, x_2, y_2]$):} Represents a keyline as two points.
    Supervision via standard \gls{oks} loss allows direct regression of the support endpoints ($\boldsymbol{e}_1$, $\boldsymbol{e}_2$). Supervision via \gls{okls} loss optimizes the induced line, letting the model predict two arbitrary points on the line.
    \item \textbf{Normal Point ($[x_n, y_n]$):} Regresses the orthogonal projection of the output cell center $\boldsymbol{c}_h$ onto the infinite keyline.
    Supervision is possible via either \gls{oks} or \gls{okls}. 
    The representation is undefined at $[0, 0]$ (lines passing through $\boldsymbol{c}_h$) and exhibits increased angular sensitivity near this singularity.
\end{itemize}

All ablation variants are trained without the keypoint output to isolate parameterization and loss effects.
For \gls{okls} application, predicted parameters are transformed to the continuous Hesse normal form.
All models are evaluated on the \gls{okls} metric.

\subsection{Unstructured line detection baselines}
\label{sec:baselines}

We benchmark against M-LSD~\citep{gu2022towards} and LINEA~\citep{janampa2025linea} as representative general-purpose line-segment detectors by training them on the same runway data. 
Both start from public pretrained checkpoints with default training settings.

Because unstructured detectors output unordered image-level lines, they cannot natively be evaluated directly using the \gls{okls}-based $\text{AP}_{\text{KL}}$ and $\text{AR}_{\text{KL}}$ metrics. 
We therefore evaluate them using an \emph{oracle best-case matching protocol} (detailed in \cref{supp:general_line_eval}). 
This protocol uses ground-truth object annotations to independently select the highest-similarity predicted line for each semantic keyline slot.
Using this matching, we can apply the standard COCO evaluation protocol.
We compare model latencies by measuring average inference time at batch size 1 on an NVIDIA Tesla T4 GPU.

\subsection{Training and implementation details}

We instantiate our framework on the YOLO-NAS-Pose Nano architecture using the SuperGradients library~\citep{supergradients}, selected for its low inference latency and decoupled multi-scale head design typical of modern YOLO models.
The network operates at an input resolution of $1024 \times 1024$ pixels across three detection heads with downsampling strides $8$, $16$, and $32$.
To evaluate model scaling, we additionally train a YOLO-NAS-Pose Medium variant, replacing the model architecture but keeping the keyline prediction and training details equivalent.

For runway perception, the model predicts $K=4$ keypoints (runway corners) and $L=5$ semantically ordered keylines: the left, top, right, and bottom boundary edges, alongside the runway centerline. 
We include the centerline because it provides a critical visual reference for estimating the aircraft's lateral alignment, particularly during short final approach (see \cref{fig:partially_visible_runway}).
Models predicting only keylines or keypoints contain $9.89$\,M parameters.
For multi-task prediction, the fused-head variant adds negligible additional model parameters, while the split-head variant increases the parameter count to $11.28$\,M ($+14.1\%$). 

All models are initialized with COCO Pose~\citep{lin2014microsoft} pretrained weights~\citep{supergradients} and trained for $80$ epochs using AdamW with standard PyTorch parameters, an effective batch size of $96$, weight decay $10^{-4}$, and a peak learning rate of $2 \times 10^{-3}$ with cosine decay.
The multi-task loss weight is $\alpha = 0.5$.
\section{Results}
\label{sec:results}

We evaluate the keyline parameterization, supervision, and joint prediction architecture, and compare our method with unstructured line-segment detectors.

\subsection{Representation and supervision ablation}
\label{sec:representation_study}

\begin{figure*}
  \centering
  \resizebox{\textwidth}{!}{\includegraphics{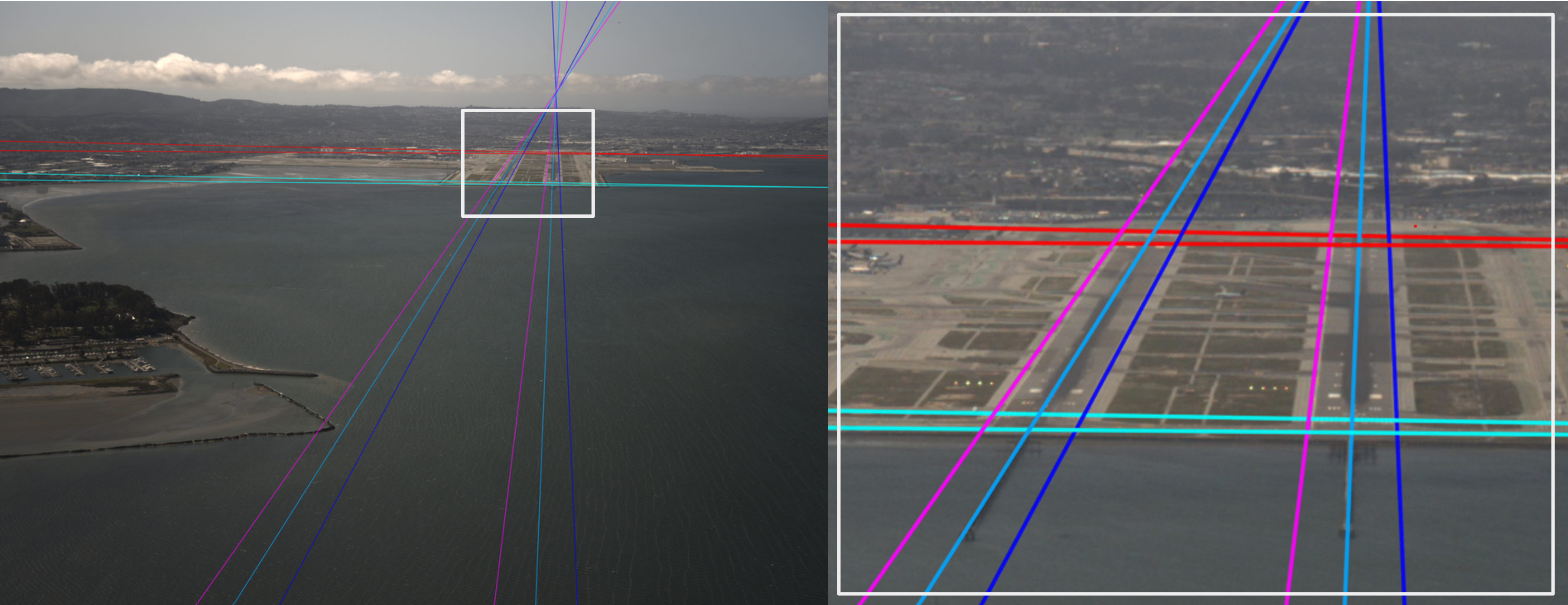}}
  \caption{Keyline predictions on a two-runway scene (San Francisco International Airport). The zoomed view (right) shows all five predicted keylines per runway with accurate geometric alignment, illustrating the multi-instance prediction capabilities of our method.}
  \label{fig:qualitative_keylines}
\end{figure*}

\begin{figure}[htbp]
  \centering
  \resizebox{\linewidth}{!}{\includegraphics{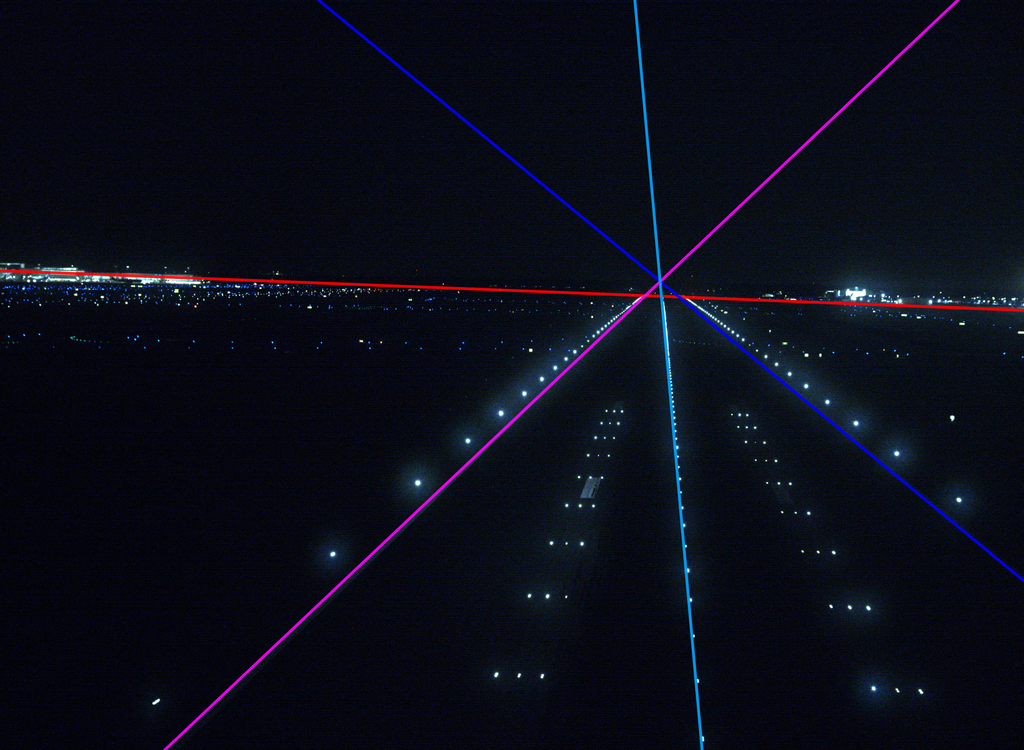}}
  \caption{Keyline predictions under nighttime conditions with truncated runway edges. The model recovers all visible keylines, although night scenes are underrepresented in training.}
  \label{fig:night_keylines}
\end{figure}

\Cref{tab:representation} benchmarks the line parameterizations and supervision variants from \cref{sec:alt_rep}. 
The continuous Hesse-normal representation $[\rho, \cos\theta, \sin\theta]$ paired with \gls{okls} supervision achieves the best performance. 
For both endpoints and normal points, supervising via \gls{okls} substantially outperforms coordinate-level \gls{oks}.
While expected given its alignment with the target metric, this confirms that \gls{okls} is generally well-suited as a loss function in YOLO and supervising decoded keyline geometry directly is more effective for keyline regression than optimizing raw parameter coordinates, whose accuracy does not directly correspond to the accuracy of resulting keylines. 
For normal points, OKS does not reflect the increased angular sensitivity near the representation's singularity; for endpoints, it unnecessarily penalizes placement along the keyline.
Qualitative predictions are illustrated in \cref{fig:qualitative_keylines,fig:night_keylines}.

\begin{table}[t]
\centering
\caption{Ablation results comparing proposed and alternative line representations and loss functions. Values are mean (sample standard deviation) over three training runs with different seeds.}
\label{tab:representation}
\small
\begin{tabular}{lccc}
\toprule
Representation & Loss & $\text{AP}_{\text{KL}} \uparrow$ & $\text{AR}_{\text{KL}} \uparrow$ \\
\midrule
\textbf{Continuous Hesse Normal} & OKLS & \textbf{55.7} (0.8) & \textbf{61.7} (0.7) \\
Polar Hesse Normal & OKLS & 53.0 (1.0) & 59.5 (0.4) \\
Endpoints & OKLS & 52.3 (1.2) & 58.9 (1.4)\\
Endpoints & OKS & 47.5 (0.8) & 53.8 (1.0) \\
Normal Point & OKLS & 52.6 (1.0) & 59.2 (0.8)\\
Normal Point & OKS & 46.5 (0.5) & 52.1 (1.3) \\
\bottomrule
\end{tabular}
\end{table}

\subsection{Multi-task prediction}
\label{sec:multi_task}

\begin{table}[!b]
\centering
\caption{Joint keypoint and keyline prediction results (mean with sample standard deviation over three training runs).  ``Original'' is the YOLO-NAS-Pose baseline trained only for runway-corner keypoints. ``Fused'' predicts keypoints and keylines through a shared geometric head, whereas ``Split'' uses a separate geometric branch for keyline regression (\cref{sec:model_architecture}). The joint models achieve comparable keypoint AP/AR to the keypoint-only baseline. While Fused and Split have comparable keyline performance, Fused incurs lower latency.}
\label{tab:multi_task_results}
\small
\resizebox{\columnwidth}{!}{%
\begin{tabular}{lccccc}
\toprule
Arch. & $\text{AP}_{\text{KL}} \uparrow$ & $\text{AR}_{\text{KL}} \uparrow$ & $\text{AP}_{\text{KP}} \uparrow$ & $\text{AR}_{\text{KP}} \uparrow$ & Lat. [ms] $\downarrow$ \\
\midrule
Original & N/A & N/A & 48.4 (0.5) & 51.5 (0.4) & 27.3 (0.3) \\
Fused   & \textbf{52.3} (0.9) & 59.1 (0.9) & 48.7 (0.8) & 52.5 (0.2) & 27.4 (0.2) \\
Split   & 52.0 (1.4) & \textbf{59.3} (1.2) & \textbf{49.8} (0.2) & \textbf{52.8} (0.3) & 29.2 (0.2) \\
\bottomrule
\end{tabular}%
}
\end{table}

With the parameterization and loss fixed, we now evaluate joint keypoint and keyline prediction.
Unlike in \cref{tab:representation}, the fused- and split-head variants predict both feature types.
As shown in \cref{tab:multi_task_results}, the fused-head variant preserves keypoint accuracy within run-to-run variation, although model capacity is shared between both feature types, while the higher-capacity split-head variant slightly increases average keypoint precision. 
Keyline prediction performance is comparable between the two multi-task integrations, although both exhibit a modest drop of 3.4 (resp. 3.7) $\text{AP}_{\text{KL}}$ points relative to the keyline-only model in \cref{tab:representation}.
A likely contributing factor is the constant multi-task loss weight $\alpha = 0.5$. Since in this dataset the number of visible keylines exceeds the number of visible keypoints, this results in larger gradients on individual keypoints compared to keylines.
This balances both tasks irrespective of feature count, as is standard in YOLO multi-task learning.
The fused-head variant does not meaningfully increase model latency, while the split-head variant increases latency by 1.9 ms (7.0\%), confirming that both keypoints and keylines can be modeled with a shared prediction head.

\paragraph{PnPL constraint availability.}
As shown in \cref{tab:pnpl_constraint_availability}, adding keylines substantially increases the proportion of images with at least three valid runway features. 
This is consistent with truncation during final approach: when the two nearest corners leave the image (see \cref{fig:night_keylines}), at most two corner keypoints remain, whereas four
keylines may still be observable as continuous edge constraints.
In the test dataset, $77.7\,\%$ of the images contain at least three visible keypoints, upper-bounding $\mathcal{A}_{\text{PnPL}}$ for the keypoints-only model at that value, whereas $100\,\%$ of the images contain at least three visible keylines.

To control for the different number of feature slots between keypoints and keylines, we compute $\mathcal{A}_{\mathrm{PnPL},4L}$ as the constraint availability excluding the centerline. 
The lines-only and joint model retain $97.1\%$ and $99.0\%$ availability, respectively, compared with $71.3\%$ for the four-keypoint model. 
Thus, the gain is not explained solely by the original five-versus-four feature count.

\begin{table}[t]
\centering
\caption{PnPL constraint availability ($\mathcal{A}_{\text{PnPL}}$) across model configurations, measuring the detection of at least three features (which is a necessary condition for pose estimation). $\mathcal{A}_{\text{PnPL}}$ uses all predicted features, while $\mathcal{A}_{\text{PnPL},4L}$ excludes the centerline. ``Ours'' refers to the proposed fused head model.}
\label{tab:pnpl_constraint_availability}
\small
\begin{tabular}{lcc}
\toprule
Model Configuration & $\mathcal{A}_{\text{PnPL}} \uparrow$ & $\mathcal{A}_{\text{PnPL},4L} \uparrow$ \\
\midrule
\textbf{Ours} (Keypoints + Keylines) & \textbf{99.3\%} & \textbf{99.0\%} \\
Lines-only & 98.8\% & 97.1\% \\
Original (Keypoints-only) & 71.3\% & N/A \\
\bottomrule
\end{tabular}
\end{table}

\subsection{Comparison with unstructured detectors}
\label{sec:unstructured_comp}

\begin{figure*}[ht]
  \centering
  \includegraphics[width=\linewidth]{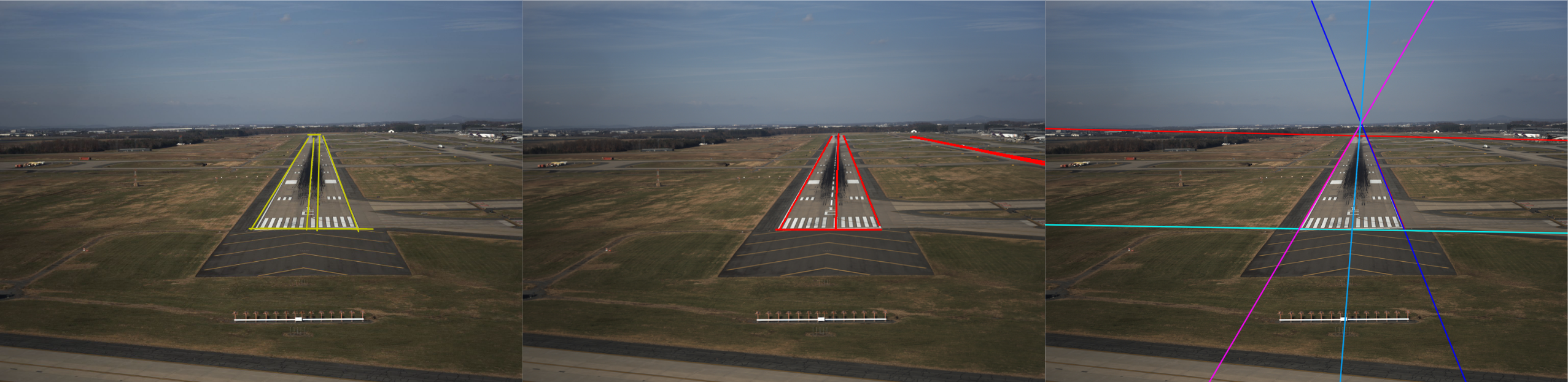}
  \caption{Common failure modes of unstructured line detectors. M-LSD (left) produces a duplicate centerline prediction. LINEA (middle) misses the far runway edge, but produces additional unwanted predictions for a taxiway with strong linear features on the right side. Neither model predicts a fixed semantic keyline set; our method (right) correctly produces predictions for the five defined runway lines.}
  \label{fig:linea_pathologies}
\end{figure*}

We compare our keyline detector with two general line-segment detectors (\cref{sec:baselines}) using a best-case oracle matching protocol.
As shown in \cref{tab:model_comparison}, our method achieves the highest runway keyline AP and AR. 
Although slower, LINEA Medium attains competitive recall ($\mathrm{AR}_{\mathrm{KL}}=58.8$), showing that unstructured detectors can recover many runway keylines. 
Its lower average precision ($\mathrm{AP}_{\mathrm{KL}}=42.4$), however, highlights fundamental limitations of evaluating unordered segments at the instance level.
Because our oracle matching evaluates complete object hypotheses, missing even a single keyline severely penalizes the aggregated instance similarity.
As illustrated in \cref{fig:linea_pathologies}, unstructured baselines lack keyline-specific semantic priors: there is no mechanism to predict exactly one of each type of keyline per object.
Hence, they may assign lower confidence to less salient runway keylines, such as the distant threshold edge, compared to more salient irrelevant edges.
Scaling our proposed model from Nano to Medium improves keyline detection, at the cost of increased latency.

\begin{table}
    \centering
    \caption{Comparison of our proposed method (keylines-only) with general line detection models evaluated on semantic keyline benchmark metrics. Best values are bold. Latencies as in \cref{sec:baselines}.}
    \label{tab:model_comparison}
    \small
    \setlength{\tabcolsep}{5pt}
    \begin{tabular}{lcccc}
    \toprule
    Model & $\text{AP}_{\text{KL}} \uparrow$ & $\text{AR}_{\text{KL}} \uparrow$ & Latency [ms] $\downarrow$  \\
    \midrule
    M-LSD Tiny & 10.7 & 26.3 & \textbf{5.6} \\
    M-LSD Large & 11.0 & 26.7 & 34.6  \\
    LINEA Small  & 39.6 & 58.7 & 43.4  \\
    LINEA Medium & 42.4 & 58.8 & 58.9  \\
    Ours Nano & 55.7 & 61.7 & 27.4 \\
    \textbf{Ours Medium} & \textbf{59.1} & \textbf{65.7} & 81.1 \\
    \bottomrule
    \end{tabular}
\end{table}

\subsection{Scope and limitations}
Object keyline detection assumes that instances of a category possess a fixed, semantically ordered set of approximately linear features. 
Runways satisfy this assumption particularly well because their edges and centerline have stable semantic identities.
Although the proposed formulation does not rely on runway-specific geometry, all empirical results in this paper are obtained on runways. 
Other potential applications include document scanning and shipping-container or parcel perception for automated logistics. 
However, we do not evaluate or claim transfer to these domains. 
The formulation may be less suitable for deformable objects, categories with variable edge topology, curved boundaries, or features whose semantic identities are ambiguous across instances.

This study evaluates keyline regression in the image domain rather than its effect on downstream pose estimation. Consequently, the results do not establish that predicted keylines improve aircraft pose accuracy. 
Evaluating their impact on 6-DoF pose error within a complete PnPL pipeline, including comparisons with keypoint-only and alternative perception pipelines, remains an important direction for future system-level evaluation.
Additionally, we do not enforce geometric constraints on keylines, such as a shared vanishing point for the longitudinal runway edges and centerline, to avoid structural error coupling across features and maintain a generic problem formulation.
For system integration, including such constraints should be evaluated for their effect on downstream pose accuracy.
\section{Conclusion}
\label{sec:conclusion}

In this work, we introduced object keyline detection, the task of predicting a fixed, semantically ordered set of infinite lines per object instance.
To solve this task efficiently, we extended the YOLO framework to predict and process keylines and proposed \gls{okls} as an evaluation metric, target assignment signal, and regression loss. 
We instantiated and evaluated our approach for runway perception in vision-based aircraft landing, a safety-critical application where runway edges often remain visible even when runway keypoints are occluded or truncated.
A constraint availability study showed that keyline detection improved feature availability for downstream aircraft pose estimation compared to predicting runway corner keypoints alone.

A line parameterization study established that the Hesse normal form optimized with \gls{okls} outperforms point-based and scalar-angle alternatives.
Integrating our keyline-detection approach into existing YOLO keypoint heads does not require substantial architectural changes: a fused head predicts keypoints and keylines jointly with negligible latency overhead, and multi-task training does not degrade keypoint accuracy, at a modest cost in keyline precision.

Against general-purpose line-segment detectors, the instance-centric approach achieves higher average precision under oracle matching by naturally enforcing structured slot assignment and penalizing missing semantic edges at the instance level.
Two directions follow naturally: integrating predicted keylines into \gls{pnpl} pose estimation to evaluate their contribution to aircraft pose accuracy during final approach, and validating the formulation across object categories beyond runways.

\section*{Acknowledgements}
This work was supported by Airbus Central R\&T.
The authors thank Thierry Sammour, Jérémie Chheang, and Pierre-Alexandre Murena for constructive discussions, operational support, and valuable guidance.
\FloatBarrier
{
    \small
    \bibliographystyle{ieeenat_fullname}
    \bibliography{main}
}

\clearpage
\appendix
\setcounter{figure}{0}
\setcounter{table}{0}
\setcounter{equation}{0}
\renewcommand{\thefigure}{\Alph{section}\arabic{figure}}
\renewcommand{\thetable}{\Alph{section}\arabic{table}}
\renewcommand{\theequation}{\Alph{section}\arabic{equation}}

\begin{strip}
\begin{center}
{\large\textbf{Appendix}}
\end{center}
\end{strip}

\section{Oracle evaluation protocol for general line detectors}
\label{supp:general_line_eval}

Unstructured line segment detectors predict an unorganized set of $P$ line segments $\hat{\mathcal{L}}_{\text{img}} = \{(\hat{\boldsymbol{l}}_p, \hat{s}_p)\}_{p=1}^P$, where $\hat{\boldsymbol{l}}_p$ represents the line geometry and $\hat{s}_p \in [0, 1]$ denotes its confidence score. Unlike keyline detectors, these models output no instance-level bounding boxes, category labels, or fixed semantic slot ordering.

To evaluate general line detectors under our instance-centric keyline detection benchmark ($\mathrm{AP}_{\mathrm{KL}}$ and $\mathrm{AR}_{\mathrm{KL}}$), we construct an \textit{oracle best-case object hypothesis} for each ground-truth object instance $m$.

\paragraph{Ground-truth definition.}
Each ground-truth object instance $m$ consists of $L$ semantic keylines
\begin{equation*}
    \mathcal{K}^\mathrm{lines}_m =\left(\left(\boldsymbol{l}_{m,l}, v_{m,l}^{\mathrm{kl}}\right) \right)_{l=1}^{L},
\end{equation*}
where each keyline $\boldsymbol{l}_{m,l}$ is defined by its finite support endpoints $(\boldsymbol{e}_{m,l,1}, \boldsymbol{e}_{m,l,2})$ and associated with visibility $v_{m,l}^{\mathrm{kl}}$ (Sec. 3.1 main paper), alongside an object area $s_m^2$ derived from the instance bounding box area.

\paragraph{Best-case line slot matching.}
For a given ground-truth object $m$ and each visible semantic line slot $l \in \mathcal V^{\mathrm{kl}}$ (Sec. 3.3 main paper), we evaluate all predicted lines $\hat{\boldsymbol{l}}_p \in \hat{\mathcal{L}}_{\text{img}}$ using the individual OKLS metric with tolerance $\kappa = 0.035$:
\begin{equation*}
\text{OKLS}(\hat{\boldsymbol{l}}_p, \boldsymbol{l}_{m,l}) := \exp \left( -\frac{p_{p,1}^2 + p_{p,2}^2}{2 \, s_m^2 \, \kappa^2} \right),
\end{equation*}
where $p_{p,1}$ and $p_{p,2}$ are the perpendicular distances from the support endpoints $(\boldsymbol{e}_{m,l,1}, \boldsymbol{e}_{m,l,2})$ to the infinite extension of predicted line $\hat{\boldsymbol{l}}_p$.

For each slot $l$, we select the optimal predicted line index $\hat{p}_m(l)$ that maximizes similarity:
\begin{equation*}
\hat{p}_m(l) = \arg\max_{p \in \{1, \dots, P\}} \text{OKLS}(\hat{\boldsymbol{l}}_p, \boldsymbol{l}_{m,l}).
\end{equation*}

\paragraph{Instance aggregation and COCO ranking.}
The overall reconstructed object-level similarity score $\text{OKLS}_{\text{obj}, m}$ and its aggregated object confidence score $\hat{q}_m$ are defined as the mean over all visible slots:
\begin{equation*}
\text{OKLS}_{\text{obj}, m} = \frac{1}{|\mathcal V^{\mathrm{kl}}|} \sum_{l \in \mathcal V^{\mathrm{kl}}} \text{OKLS}(\hat{\boldsymbol{l}}_{\hat{p}_m(l)}, \boldsymbol{l}_{m,l}),
\end{equation*}
\begin{equation*}
    \hat{q}_m = \frac{1}{|\mathcal V^{\mathrm{kl}}|}  \sum_{l \in \mathcal V^{\mathrm{kl}}} \hat{s}_{\hat{p}_m(l)}.
\end{equation*}

These reconstructed object hypotheses $(\hat{q}_m, \text{OKLS}_{\text{obj}, m})$ are ranked by confidence score $\hat{q}_m$ and evaluated using standard COCO 101-point interpolated precision-recall curves across thresholds $\tau \in \{0.50, 0.55, \dots, 0.95\}$.
If $\text{OKLS}_{\text{obj}, m} \ge \tau$, the instance prediction is registered as a true positive, and as a false positive otherwise.
Because this protocol grants unstructured detectors ground-truth-guided slot assignment and permits line reuse across slots, it represents an optimistic upper bound on their instance-level detection performance.

\section{Cell-relative keyline decoding}
\label{supp:cell_relative_decoding}

At each output grid cell, the detector initially predicts unconstrained raw line parameters
\begin{equation*}
    \tilde{\boldsymbol{l}}_l =
    [\tilde{\rho}_l, \tilde{n}_{x,l}, \tilde{n}_{y,l}] \in \mathbb{R}^3.
\end{equation*}
To achieve a valid unit normal vector, the raw directional outputs are passed through a $\tanh$ activation
and normalized:
\begin{equation*}
    \boldsymbol{u}_l =
    \begin{bmatrix}
        \tanh(\tilde{n}_{x,l})\\
        \tanh(\tilde{n}_{y,l})
    \end{bmatrix},
    \qquad
    \hat{\boldsymbol{n}}_l =
    \frac{\boldsymbol{u}_l}
         {\max(\|\boldsymbol{u}_l\|_2,\epsilon)},
\end{equation*}
where $\epsilon = 10^{-12}$ is a small constant to prevent division by zero.

The raw predicted distance $\tilde{\rho}_l$ represents the local distance between the output cell center $\mathbf{c}_h$ and the predicted line in the head's output coordinates.
We therefore first scale the predicted distances to the model-input coordinate system
\begin{equation*}
    \rho_{\mathrm{scaled},l}=s_h\tilde{\rho}_l,
\end{equation*}
with the respective head stride $s_h\in\{8,16,32\}$.
Secondly, we convert the cell-relative line
\begin{equation*}
    \hat{\boldsymbol{n}}_l^\top(\mathbf{x}-\mathbf{c}_h)
    =\rho_{\mathrm{scaled},l}
\end{equation*}
to global Hesse normal form (relative to the coordinate origin) using
\begin{equation*}
    \hat{\rho}_{\mathrm{global},l}
    =\mathbf{c}_h^\top\hat{\boldsymbol{n}}_l
     +\rho_{\mathrm{scaled},l}.
\end{equation*}
The decoded head output is therefore
\begin{equation*}
    \hat{\boldsymbol{l}}_l =
    [\hat{\rho}_{\mathrm{global},l},\hat{n}_{x,l},\hat{n}_{y,l}].
\end{equation*}
All subsequent loss, assignment, and evaluation operations use this
global representation.

\section{Inference-time post processing}
\label{supp:inference}

In YOLO, input images are resized and sometimes padded, to adhere to a standard, typically square, input resolution.
Outputs, such as keypoints and keylines, must therefore be mapped back to the original image.
Let
\begin{equation*}
    \boldsymbol{x}_{\mathrm{model}}
    =
    f_s\boldsymbol{x}_{\mathrm{orig}}+\boldsymbol{p}
\end{equation*}
denote the uniform resize and padding transformation.
To map keylines to the original coordinate system, the unit normal can be left unchanged, while the line distance is transformed to the new origin and unscaled:
\begin{equation*}
    \hat{\rho}_{\mathrm{orig},l}
    =
    \frac{
        \hat{\rho}_{\mathrm{global},l}
        -\hat{\boldsymbol{n}}_l^\top\boldsymbol{p}
    }{
        f_s
    }.
\end{equation*}
\clearpage

\begin{strip}
\section{Example approach predictions}
\label{supp:example_predictions}
\vspace{1.5ex}

\centering
\includegraphics[width=0.33\textwidth]{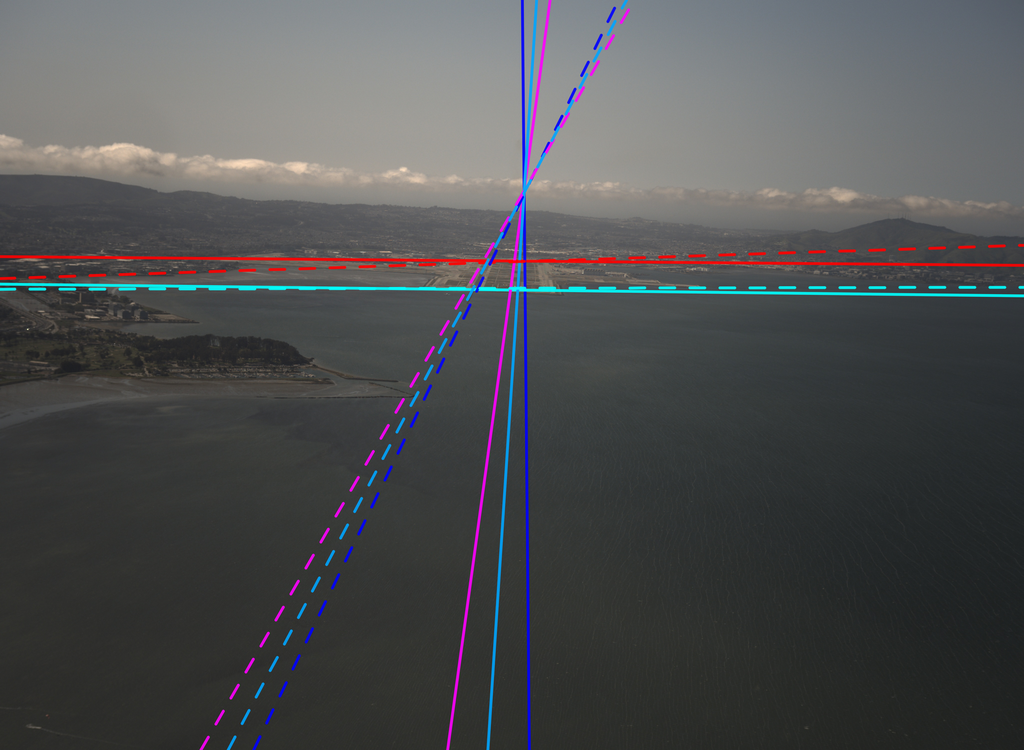}\hfill
\includegraphics[width=0.33\textwidth]{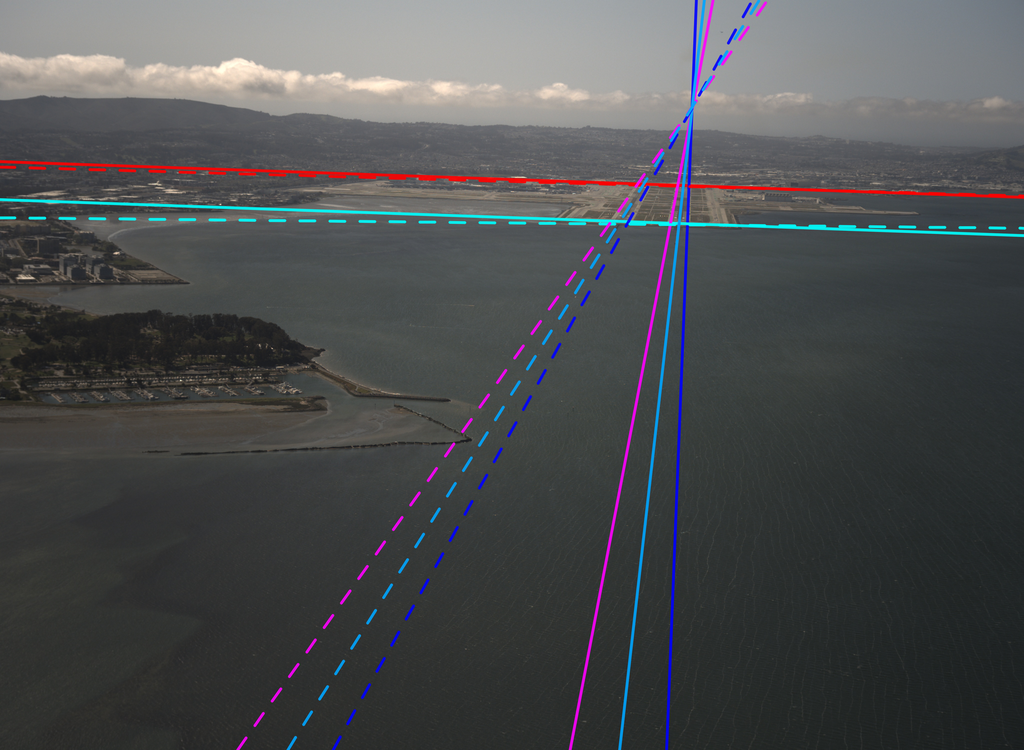}\hfill
\includegraphics[width=0.33\textwidth]{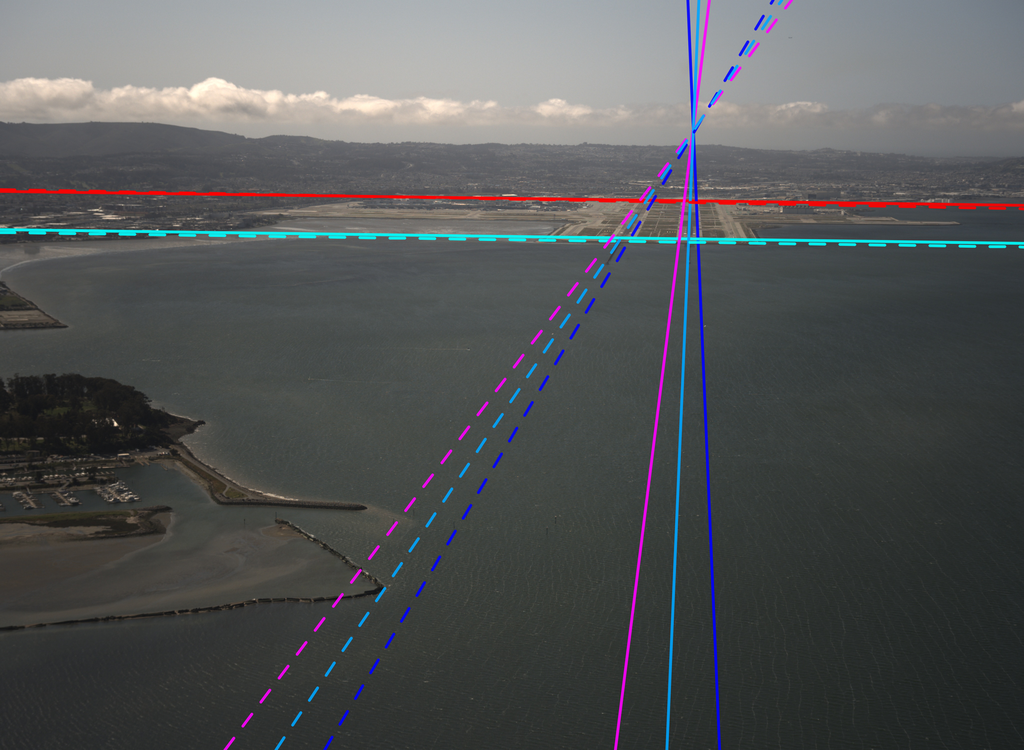}\\[0.4ex]

\includegraphics[width=0.33\textwidth]{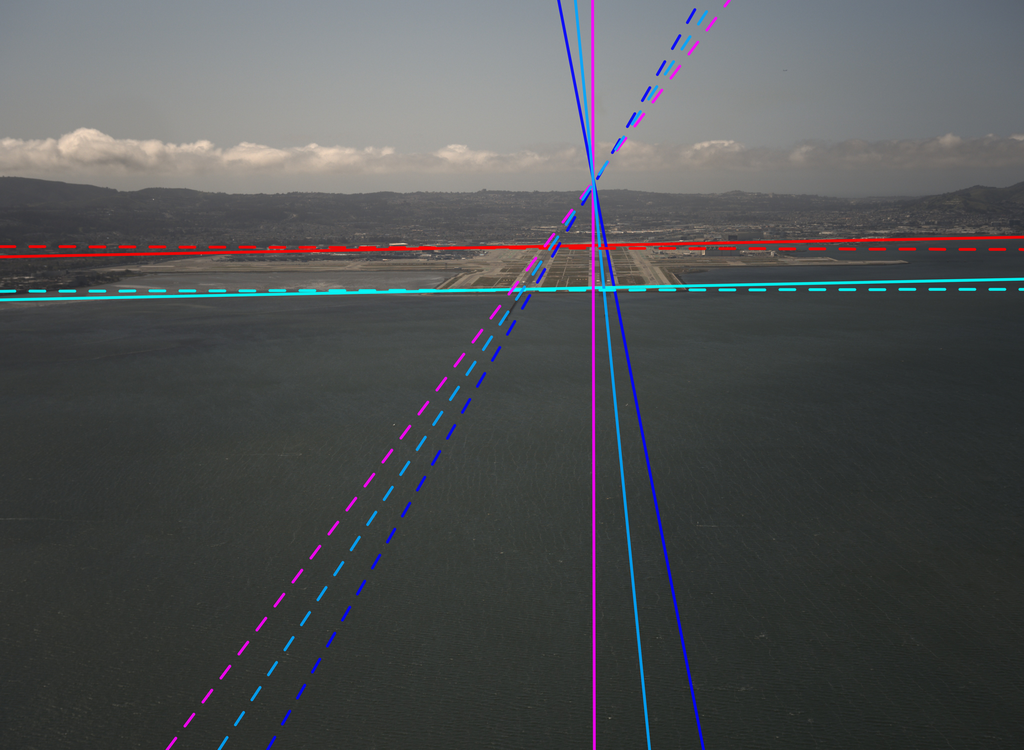}\hfill
\includegraphics[width=0.33\textwidth]{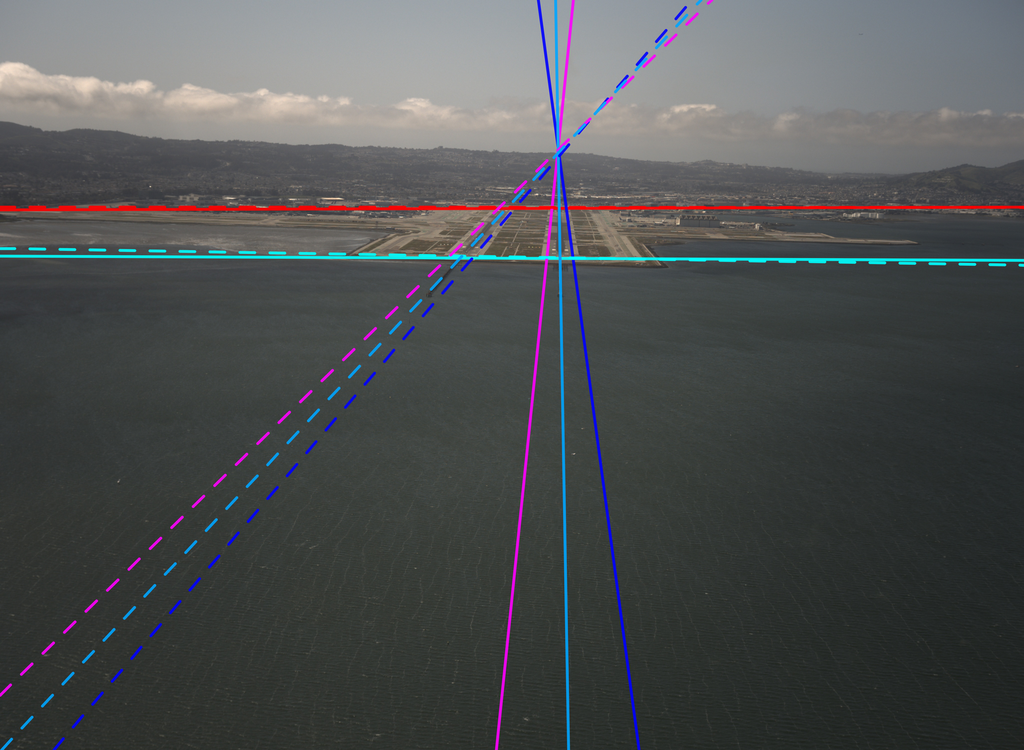}\hfill
\includegraphics[width=0.33\textwidth]{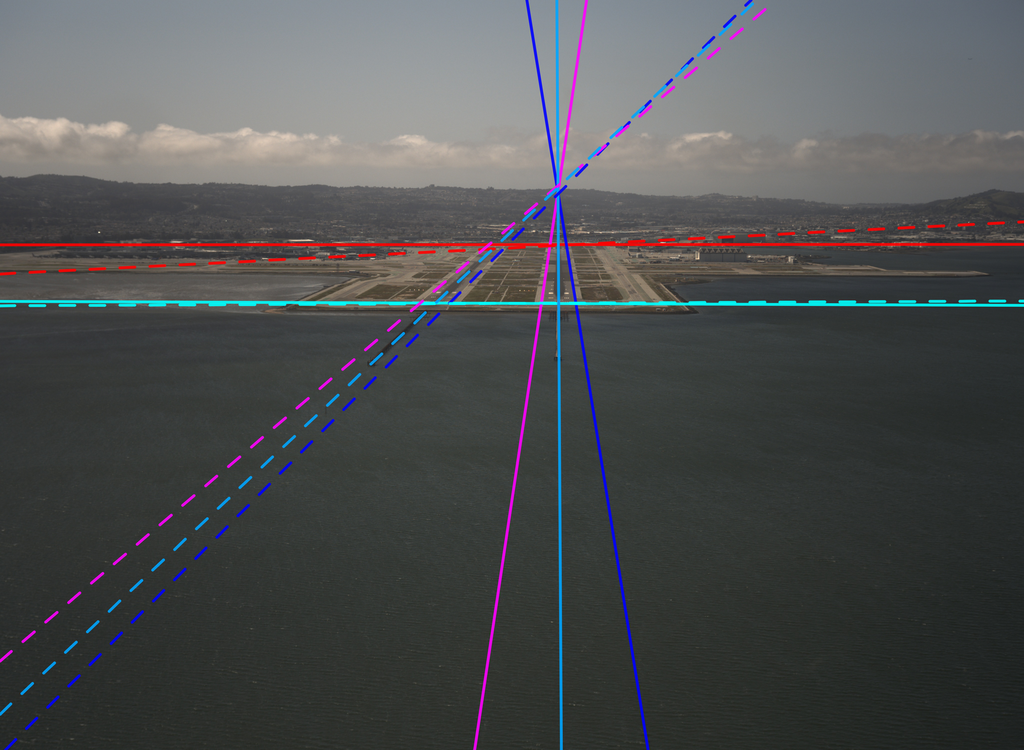}\\[0.4ex]

\includegraphics[width=0.33\textwidth]{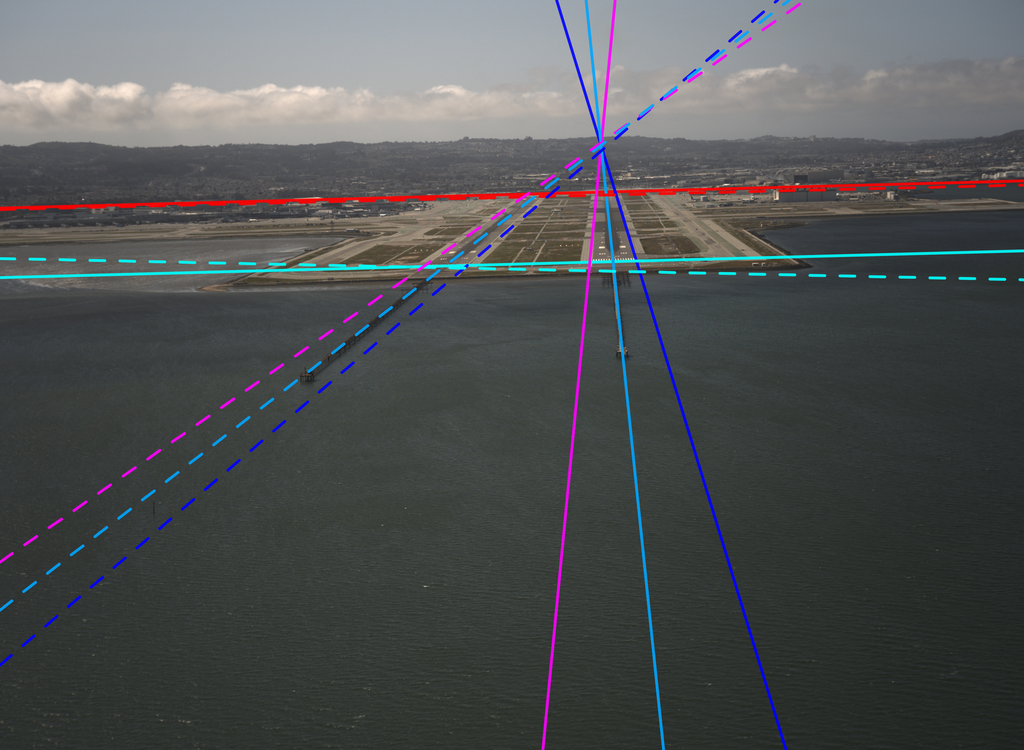}\hfill
\includegraphics[width=0.33\textwidth]{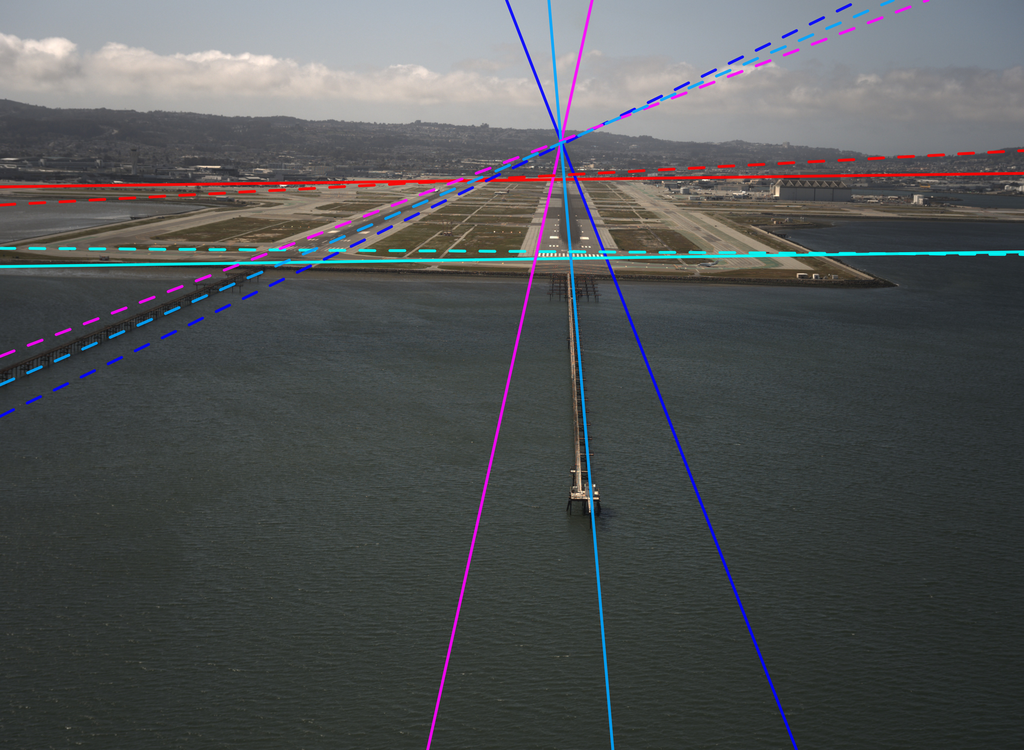}\hfill
\includegraphics[width=0.33\textwidth]{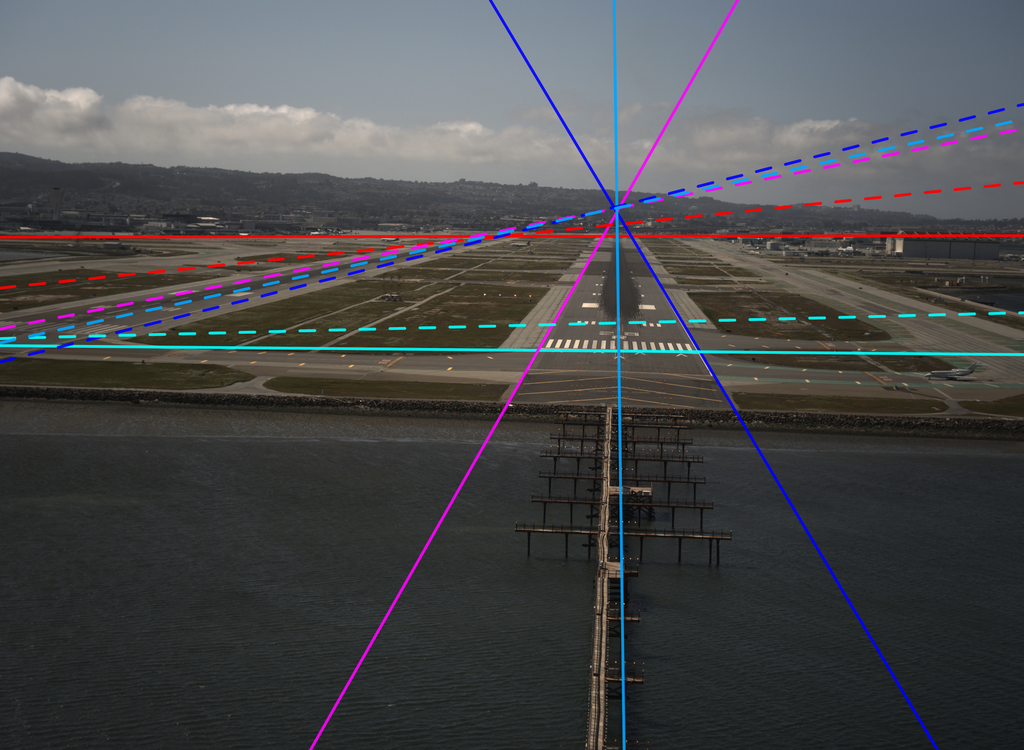}\\[0.4ex]

\includegraphics[width=0.33\textwidth]{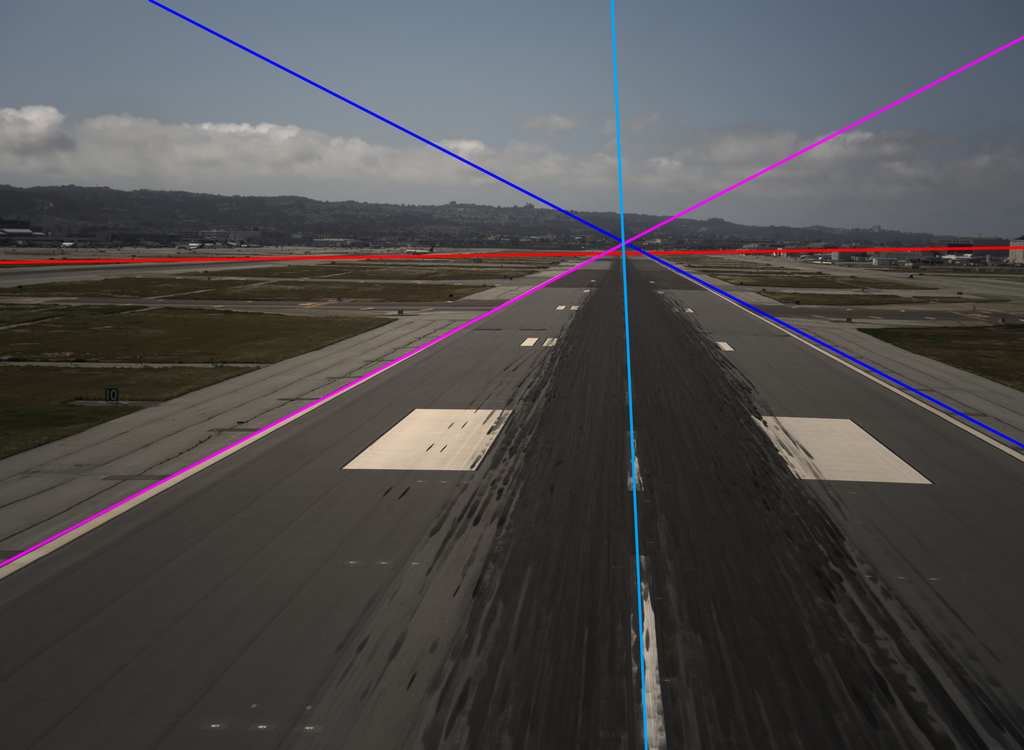}\hfill
\includegraphics[width=0.33\textwidth]{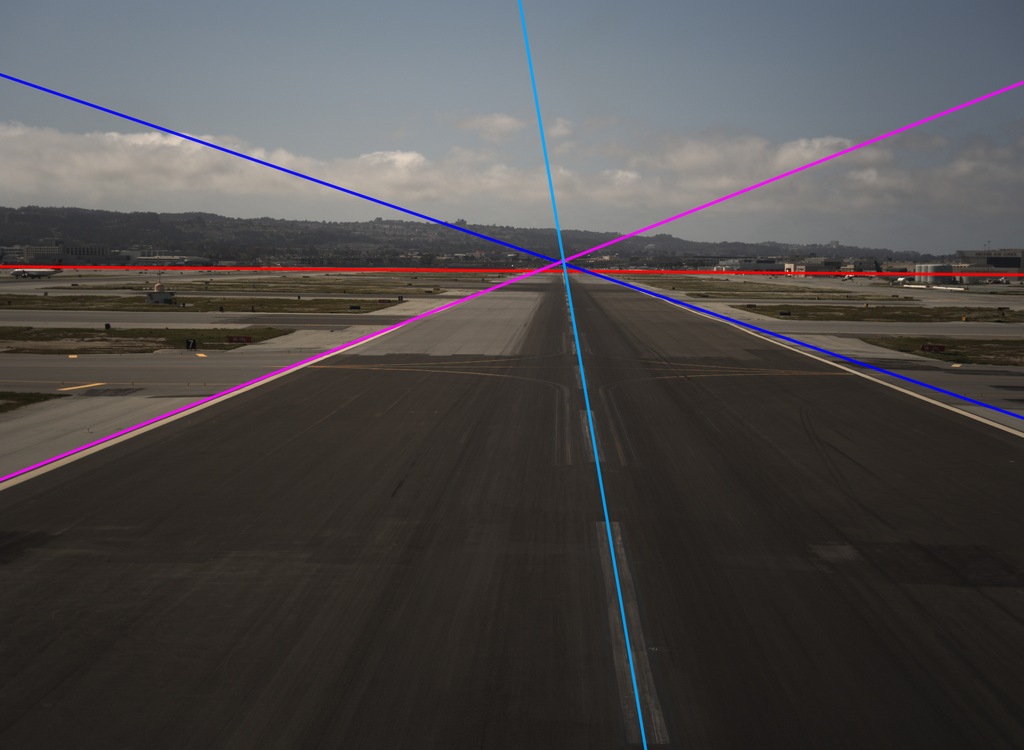}\hfill
\includegraphics[width=0.33\textwidth]{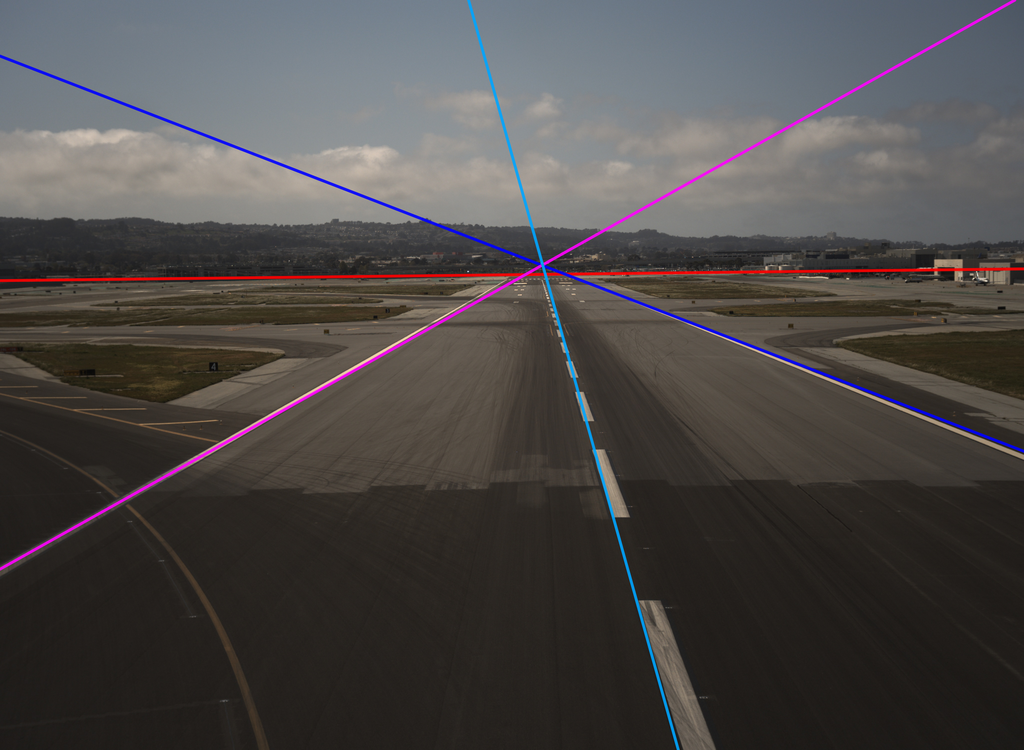}

\vspace{1ex}
\captionof{figure}{Keyline predictions recorded during an approach to San Francisco International Airport (SFO), sampled at 12-second intervals. Solid lines denote the highest-confidence predictions, while dashed lines show predictions with the second-highest confidence (when present).}
\label{fig:prediction_grid}
\end{strip}

\end{document}